\documentclass{scrartcl}

\usepackage{amsmath}
\usepackage{amssymb}
\usepackage{graphicx}
\usepackage{booktabs}
\usepackage[margin=1.0in]{geometry}

\usepackage{verbatim}
\usepackage{enumitem}
\usepackage[utf8]{inputenc}
\usepackage[english]{babel}
\usepackage{csquotes}

\usepackage{caption}
\usepackage[OT2,T1]{fontenc}
\usepackage{biblatex}
\usepackage[colorlinks=true]{hyperref}
\usepackage{cleveref}

\title{Transformers Learn the Mestre-Nagao Heuristic}
\author{Pranav Venkata Konda}
\date{}

\DeclareSymbolFont{cyrletters}{OT2}{wncyr}{m}{n}
\DeclareMathSymbol{\Sha}{\mathalpha}{cyrletters}{"58}

\begin{document}
\maketitle 
\begin{abstract}
  \small
We train a two-layer transformer encoder to classify
rational elliptic curves $E/\mathbb{Q}$ of conductor $\leq
10000$ as either rank 0 or rank 1 from the first 128
normalized Frobenius traces. We achieve >99\% accuracy on
both classes, and accuracy is essentially unchanged on test
curves with no isogeny or quadratic-twist relative in the
training set. We then apply techniques from mechanistic
interpretability such as attention analysis, linear probing,
activation patching, logit attribution, and neuron-level
circuit analysis to reverse-engineer the algorithm the
(centroid in function space) model
learned. 
We find that a sparse circuit of 20 out of 512 layer-1 MLP
neurons is sufficient for rank prediction under a 
linear probe with an AUROC of
0.992 at plateau, implementing a push-pull detector
architecture of rank-0 and rank-1 detectors
with a one-sided readout: rank-1 is signaled by
a withheld push rather than by an opposing pull.
However, we notice that the
model has sub-optimal readout problems: the model's readout
weights extract only an AUROC of 0.956 from the same
neurons, indicating a mismatch in rank-order between the
readout pathway and the discriminative circuit. 
Critically, the learned input weights of
the top discriminating neuron match the Mestre-Nagao sum
heuristic weights $\log(p)/(p\cdot \log{B})$ with a Spearman
coefficient $r = 0.997$ and Pearson coefficient $r = 0.952$:
the model has learnt a
result from analytic number theory from the Frobenius trace
data alone. 
We additionally find that all 50 independently trained
models concentrate CLS attention on prime positions at
2-50$\times$ the rate of composite positions, which is
consistent with the Euler product structure of $L(E, s)$.
The CLS embedding encodes $\log{L(E,
1)}$ with $R^2 = 0.962\pm 0.011$ across the 50 models (after
controlling for conductor). 
Activation patching analysis reveals that
attention weights are dissociated from causal information
flow. Additionally, the 50 solutions from training are
near-identical in function space (with pairwise agreement
$>98.8\%$) despite large weight space barriers.
\normalsize
\end{abstract}

\section{Introduction}

The Birch and Swinnerton-Dyer (BSD) conjecture 
\cite{BirchSwinnertonDyer1965} predicts that
the rank of the Mordell-Weil group $E(\mathbb Q)$ equals the
order of vanishing of the $L$-function $L(E, s)$ at $s=1$.
We note that $L(E, s)$ is completely determined by the
Frobenius traces $\{a_n\}$, from the Euler product 
\[ L(E, s) = \prod_{p} (1 - a_pp^{-s} + p^{1 - 2s})^{-1}. \]
In theory, the rank is readable thus from the sequence of
Frobenius traces $(a_1, a_2, \ldots)$. Detecting the
vanishing of the $L$-value $L(E, 1)$ from finitely many
terms is numerically very difficult, as the approximate
functional equation 
\[ L(E, 1) \approx 2 \sum_{n=1}^N \frac{a_n}{n} \cdot W\left
  (\frac{n}{\sqrt{N_E}}\right ) \] converges slowly
  (particularly for elliptic curves of high conductor)
  \cite{rubinstein2005computational}. 

Previous work in this area has shown that machine learning
models are capable of predicting rank from Frobenius traces
with high accuracy, such as \cite{babei2025learning},
\cite{bieri2026murmurations},
\cite{10.1016/j.jsc.2022.08.017}, and
\cite{kazalicki2023ranks}. These works show that prediction
using machine learning is feasible, but do not address
questions from mechanistic interpretability: what algorithm
does the model discover? 

We address this by using tools from mechanistic
interpretability (\cite{elhage2021mathematical}, 
\cite{nanda2023progressmeasuresgrokkingmechanistic}, 
\cite{elhage2022toymodelssuperposition}), such
as attention analysis, linear probing 
\cite{alain2018understandingintermediatelayersusing},
activation patching \cite{10.5555/3600270.3601532}, direct
logit attribution, and neuron-level circuit analysis. We
find that a transformer trained on rank prediction
independently rediscovers the Mestre-Nagao heuristic
\cite{bieri2026murmurations}, a result from classical
analytic number theory that estimates the rank, implemented
by a sparse push-pull MLP circuit. This appears to be the
first mechanistic identification of a transformer neural network 
learning a named
mathematical result from number-theoretic data without
supervision. 

\section{Background}

\subsection{L-functions, BSD, and Frobenius Traces}
Let $E/\mathbb Q$ be a rational elliptic curve with
conductor $N_E$. The \textbf{$L$-function} is 
\[ L(E, s) = \sum_{n\geq 1} \frac{a_n}{n^s} = \prod_{p \nmid
  N_E} \frac{1}{1 - a_pp^{-s} + p^{1 - 2s}} \cdot \prod_{p
  \mid N_E} \frac{1}{1 - a_pp^{-s}}. \] Here, $a_p = p+1 -
\#E(\mathbb F_p)$ for primes of good reduction (i.e. the
curve is non-singular when the coefficients are reduced
modulo $p$), and we have by the Ramanujan-Eichler-Shimura
bound that $|a_p| \leq 2\sqrt{p}$ \cite{diamond2005first}.

The Frobenius traces $a_n$ satisfy \textbf{Hecke
multiplicativity}: for $m, n$ where $\gcd{(m, n)} = 1$, we
have $a_{mn} = a_ma_n$. Thus $\{a_n\}$ is determined
completely by $\{a_p\}$ for primes $p$. 

For rank 0 elliptic curves, the BSD formula gives 
\[ L(E, 1) = \frac{\Omega_E \cdot \#\Sha(E)
  \cdot \prod_{p}c_p}{|\operatorname{Tor}(E(\mathbb Q))|^2}.
\] where:
\begin{enumerate}[label=\roman*.]
  \item $\Omega_E$ is the real period, defined as follows:
    note that every elliptic curve $E/\mathbb Q$ has a Weierstrass
    equation with integer coefficients: we have that $E$ is
    the projective curve 
    \[ y^2 = x^3 + ax + b.\] We can define the unique
    invariant differential (in the sense that it is
    translation invariant)
    \[ \omega_E = \frac{\operatorname{d}x}{2y}. \] The
    lattice of periods is then defined as the discrete 
    subgroup of $\mathbb C$ generated by integrals of the
    form 
    \[ \int_{\gamma} \omega, \] where $\gamma \in H_1(E,
    \mathbb Z)$ (note that there is an isomorphism
    $E(\mathbb C) \cong \mathbb C/\Lambda$). The real period
    $\Omega_E$ is then defined as the least positive element
    of $\Lambda \cap \mathbb R$ multiplied by the number of
    components of $E(\mathbb R)$ \cite{lmfdb}. 
  \item $\Sha(E)$ is the Tate-Shafarevich group, defined as
    follows: let $K$ be a number field, and let $G_K$ be its
    absolute Galois group. For a place $\nu$ let $K_{\nu}$
    denote the completion at $\nu$ of $K$, and let
    $G_{K_\nu}$ be the absolute Galois group of the
    completion. We define the Tate-Shafarevich group for an
    elliptic curve $E/K$ as 
    \[ \Sha(E) = \ker{\left ( H^1(G_K, E) \to \prod_{v}
    H^1(G_{K_v}, E_{K_v}) \right )}, \] where $\nu$ runs over
    all places of $K$, and $E_{K_\nu}$ denotes the base change
    of $E$ to $K_\nu$. The order of $\Sha(E)$ is conjectured
    to be finite. 
  \item $\prod_{\mathfrak p} c_{\mathfrak p}$ is the
    Tamagawa product, defined as follows: let $\mathfrak p$
    be a prime of $K$. We define the Tamagawa number
    \[ c_{\mathfrak p} = [E(K_{\mathfrak p} : E^0(K_{\mathfrak
    p}))], \] where $E^0(K_{\mathfrak p})$ is the subgroup
    of $E(K_{\mathfrak p})$ consisting of all points whose
    reduction modulo $\mathfrak p$ is smooth. If $E$ has
    good reduction at $\mathfrak p$, then $c_{\mathfrak
    p}(E) = 1$. 

    The Tamagawa product is the product of the Tamagawa
    numbers over all primes, and is a positive integer. 
  \end{enumerate}

We note importantly that for curves of rank 1, the $L$-value
at $s=1$, $L(E, 1) = 0$. 
\subsection{Mestre-Nagao Heuristic}

For a positive real bound $B$, the \textbf{Mestre-Nagao sum}
is defined as 
\[ S(E, B) = \frac{1}{\log{B}} \sum_{p < B, p \nmid N_E}
\frac{a_p\log{p}}{p} \] \cite{bieri2026murmurations}. 
For an elliptic curve of analytic rank $r_{\mathrm{an}}$,
the explicit formula for $\frac{L'(E, 1)}{L(E, 1)}$ predicts
that if $\lim\limits_{B \to \infty} S(E, B)$ exists, it
converges to $\frac{1}{2} - r_{\mathrm{an}}$
\cite{kim2021birchswinnertondyerconjecturenagaos}. Rank-0
curves thus satisfy $S(E, B) \approx \frac{1}{2}$, and
rank-1 curves satisfy $S(E, B) \approx -\frac{1}{2}$ for
large $B$. The sum separates the rank classes by a gap of 1
and serves as a heuristic predictor of rank with weights 
$w_p =\frac{\log{p}}{p}$ indexed by prime $p$. The per-prime
normalized Mestre-Nagao weights are 
\begin{equation}\label{eqn:mn}
w_p = \frac{\log{p}}{p\cdot\log{B}}.
\end{equation}
Bieri et al. in
\cite{bieri2026murmurations} show that Mestre-Nagao sums
achieve an AUROC of approximately 0.95 for rank prediction
and that CNN saliency curves qualitatively resemble $w_p$ as
a function of $p$. We establish this connection at the level
of individual neurons with Pearson $r=0.952$ with circuit
analysis methods. 

\subsection{Transformers and Mechanistic Interpretability}
A transformer encoder processes input $(x_1, \ldots, x_T)$
through $L$ layers, each with multi-head self-attention and
a position-wise MLP. For a head $h$ we have 
\[ \alpha_{ij}^{(h)} = \operatorname{softmax}\left (
    \frac{{q_i}^{(h)} \cdot
  k_j^{(h)}}{\sqrt{d_{\mathrm{head}}}} \right), \text{ and }
  z_i^{(h)} = \sum_j \alpha_{ij}^{(h)}v_j^{(h)}. \] We
  prepend a learned CLS token (introduced in
  \cite{devlin2019bert}) whose final hidden state is used
  for classification. 

  Mechanistic interpretability \cite{elhage2021mathematical}
  aims to reverse-engineer the learned algorithm through
  analyzing model internals, including analysis of attention
  weights \cite{clark-etal-2019-bert}, linear probing
  \cite{alain2018understandingintermediatelayersusing},
  activation patching \cite{10.5555/3600270.3601532}, and
  neuron-level circuit analysis
  \cite{nanda2023progressmeasuresgrokkingmechanistic}. 

\section{Experimental Setup}
\subsection{Data}

We use the Cremona database accessed through the L-functions
and Modular Forms Database (LMFDB) \cite{lmfdb}. We restrict
to curves of conductor $N_E \leq 10000$ and analytic rank in
$\{0, 1\}$, which yields 62,298 curves (30,427 rank 0 curves
and 31,871 rank 1 curves), with a stratified 80/20 split
(random seed 42). The input is 
\[ \tilde a_n = \frac{a_n}{2\sqrt{n}}, \] where $n = 1,
\ldots, N$ with $N=128$ as the primary experimental setting.
BSD-related invariants (such as $L$-values, periods,
Tamagawa numbers, $\Sha(E)$, and torsion) are fetched from
the LMFDB and computed by Dokchitser's algorithm
\cite{dokchitser2002computingspecialvaluesmotivic}. 

\subsection{Model}

We train a 2-layer transformer encoder $d_{\mathrm{model}} =
128$, with 4 heads, MLP width $4d_{\mathrm{model}} = 512$
neurons per layer, pre-norm LayerNorm \cite{ba2016layer} at
approximately 500,000 parameters. Training uses AdamW
\cite{loshchilov2019adamw}, lr $ = 3\times10^{-4}$, cosine
schedule \cite{loshchilov2017sgdr}, weight decay 0.01,
weighted cross-entropy for class imbalance, and 100 epochs.
We train 50 independent models with random seeds 8-57. 

The model achieves $98.7\pm 0.02\%$ accuracy on rank 0
curves and $99.5\pm0.2\%$ on rank 1 curves. This exceeds 
the Mestre-Nagao partial sum baseline AUROC of 0.95, and the
naive partial sum 
\[ S_N = 2 \sum_{n=1}^N \frac{a_n}{n}, \] which has an AUROC
of 0.93. 

Recently, Babei, Shah, and Kebe
\cite{babei2026twistclassredundancydrives} showed that for
the related task of predicting a Frobenius trace from nearby
traces, much of the reported model performance is
attributable to quadratic-twist redundancy in the dataset
as twist-classes share trace magnitudes: an explicit
twist-matching baseline substantially outperforms the
trained transformers. We show that this situation does not
apply in the case of rank prediction. 

We note that the dataset contains all curves of conductor
 $\leq 10^4$ rather than one representative per isogeny
class, so isogenous curves (which have the same trace
sequences and identical ranks) may appear in both training
and test set data. Following the discussion of
twist-redundancy analysis in
\cite{babei2026twistclassredundancydrives}, we partition the
test set into three slices: curves with exact isogeny
duplicates in the training set (60.4\% of curves), curves
with no exact duplicate with quadratic-twist proxy key (this
refers to the absolute traces $|a_p|$ at the eight largest
primes $p \leq 127$, following \cite{babei2026twistclassredundancydrives})
matching a training curve (23.5\%), and curves with neither
(16.1\%). 

Our representative model (the centroid model in function
space, see \cref{subsec:function_space}) achieves 99.7\%, 98.9\%,
and 98.9\% accuracy on these slices respectively, with AUROC
$\geq 0.999$ on each. Performance is thus essentially
unchanged on curves about which the training set carries no
twist-class information. A twist-lookup baseline that
predicts the majority rank among a test curve's twist proxy
mates in training performs at 50.4\% accuracy on the twist
slice. Rank, unlike trace magnitude, is not recoverable from
twist-class membership. This implies the model's accuracy is
not derived from isogeny or twist retrieval, consistent with
the parametric Mestre-Nagao mechanism in \cref{sec:mn_sums},
and in contrast to the trace-prediction setting of
\cite{babei2026twistclassredundancydrives}. 
\section{Solution Space Geometry}

\subsection{Weight Space}\label{sec:weight_space}
We compute the pairwise loss barriers by linear mode
connectivity following 
\cite{goodfellow2015qualitativelycharacterizingneuralnetwork}
and \cite{garipov2018losssurfacesmodeconnectivity}.

For each pair
$(i, j)$ of models, we interpolate $\theta(\alpha) =
\alpha\theta_i + (1 - \alpha)\theta_j$ over a uniform grid
of 21 values $\alpha \in [0, 1]$ and define 
$\operatorname{barrier}(i, j) = 1 -
\min_{\alpha}{\operatorname{acc}(\theta(\alpha))}$. This
agrees with the endpoint-relative barrier up to
approximately 0.01 as every endpoint model achieves an
accuracy of approximately 0.99. Across the 1225 pairs,
barriers range from 0.18 to 0.89 with a mean of 0.556: no
pair of solutions is linearly mode-connected in raw weight
space.

These interpolations are performed without permutation or
rescaling alignment of neurons. Large raw-weight
barriers between functionally equivalent models are expected
under the parameter-space symmetries of the architecture
(see \cite{ainsworth2023gitrebasinmergingmodels} and
\cite{entezari2022rolepermutationinvariancelinear}), which
is confirmed strongly by the function space analysis in 
\cref{subsec:function_space}. 

We apply metric MDS to the pairwise barrier matrix treated
as a dissimilarity matrix to classify the geometry of
the weight space using Kruskal stress-1. The stress declines
with embedding dimension until $\approx d = 15-20$, and
plateaus at $\approx 0.15$ and improves no further even at
$d=49$. This implies the barrier geometry admits no
low-dimensional Euclidean structure, and is not
approximately Euclidean at any dimension. This is consistent
with loss barriers not satisfying metric axioms. 

\subsection{Function Space}\label{subsec:function_space}
\begin{figure}[htpb]
  \includegraphics[scale=0.5]{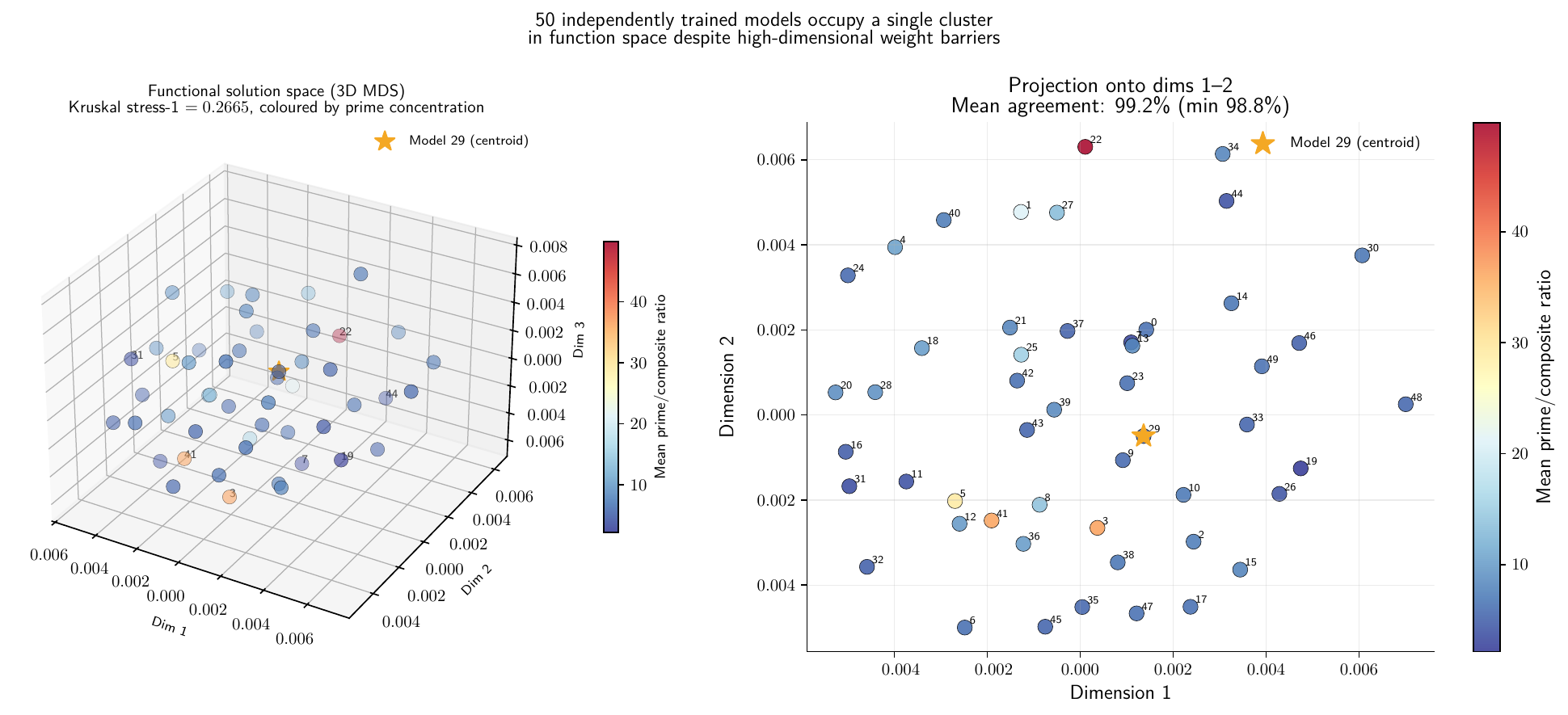}
  \caption{The left plot depicts the functional solution
  space projected onto 3 dimensions, with solutions colored
by their attention to primes for the purpose of function
space visualization. Refer to \cref{subsec:function_space}
for the stress interpretation. The right plot projects onto
dimensions 1 and 2. Notice the clustering of the solutions.}
  \centering
  \label{figs:centroid_model}
\end{figure}

We compute functional similarity of the models by Hamming
agreement on their test set predictions. The mean pairwise
agreement across all 50 models was $99.2\%$ with a standard
deviation of $0.11\%$, and all pairs of models had agreement
above $98.8\%$. Refer to \cref{figs:centroid_model} for a
visualization of the function dissimilarities $(1 -
\mathrm{agreement})$ by 3D MDS. The embedding's Kruskal
stress-1 of 0.27 reflects the near-equidistant character of
the residual disagreements, as dissimilarities range from
0.005-0.012 and near-equidistant point sets do not embed
in low dimension. This is expected from a single shared
function being perturbed by small independent per-model 
errors rather than of multiple functional clusters. 
All models are effectively
computing the same function, across many different
parameterizations (hence the large barriers observed in
weight space). We attempted to characterize function space
by various metrics, but note that prime concentration
strength does \emph{not} organize the functional space. 

With the above results in mind, the centroid model in
functional space was chosen as a representative model for
subsequent analysis (although checks of the solution types
of all models revealed that their broad structure, as
expected, was the same). 

\section{Attention Analysis}
For each trained model, we extract the mean CLS attention
weight to each input position, averaged over the test set.
Across all 50 trained models, layer-0 attention concentrates
on prime positions. Of 200 layer-0 heads, 198 of them attend
more strongly to prime positions than composite positions,
with per-head prime/composite ratios spanning 0.65$\times$ -
128$\times$ (the extreme values reflect heads with near zero
composite attention). Per-model means over the four layer-0
heads range from 2.1$\times$ to 50$\times$ (median $
6.5\times$, and mean $9.7\times$), and every model's mean
exceeds $2\times$, implying prime concentration is universal
at the model level. The two heads weakly preferring composite
positions ($0.65\times$ and $0.92\times$, in two different
models) are offset by strong prime concentration in their
sibling heads. 

\begin{figure}[htpb]
  \includegraphics[scale=0.4]{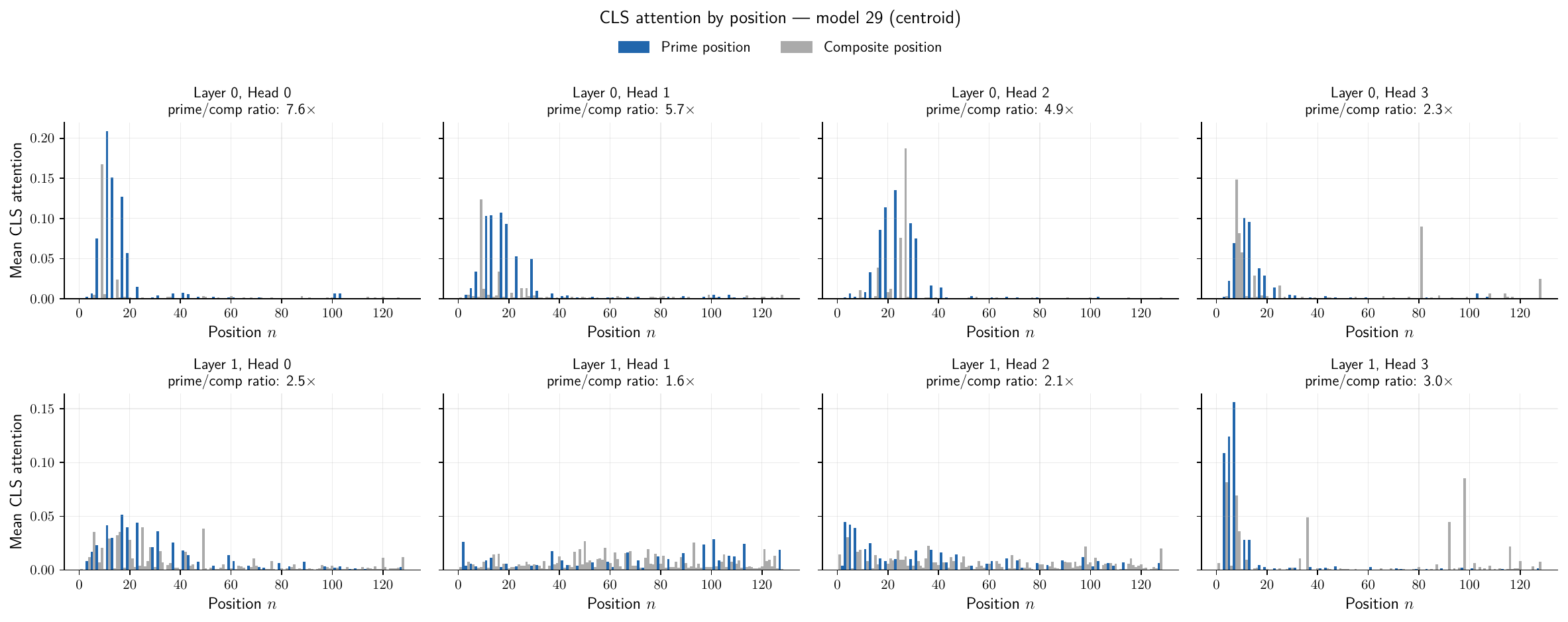}
  \caption{CLS attention weight by position for the centroid
  model. The blue bars are prime positions, grey bars are
composite positions. Note that prime positions receive
1.6-7.6$\times$ the amount of attention than composite
positions, and no head in any layer pays more attention to
composite positions.}
  \centering
  \label{fig:fig1}
\end{figure}

This prime preference is consistent
with the mathematical structure of the Euler product: since
$a_{mn} = a_ma_n$ for $\gcd(m, n) = 1$, the composite index
Frobenius traces are determined entirely by those with prime
indices, and the model learns to exploit this arithmetic
structure. Refer to \cref{fig:fig1} for the attention
distribution for the centroid model. 

\begin{figure}[htpb]
  \includegraphics[scale=0.5]{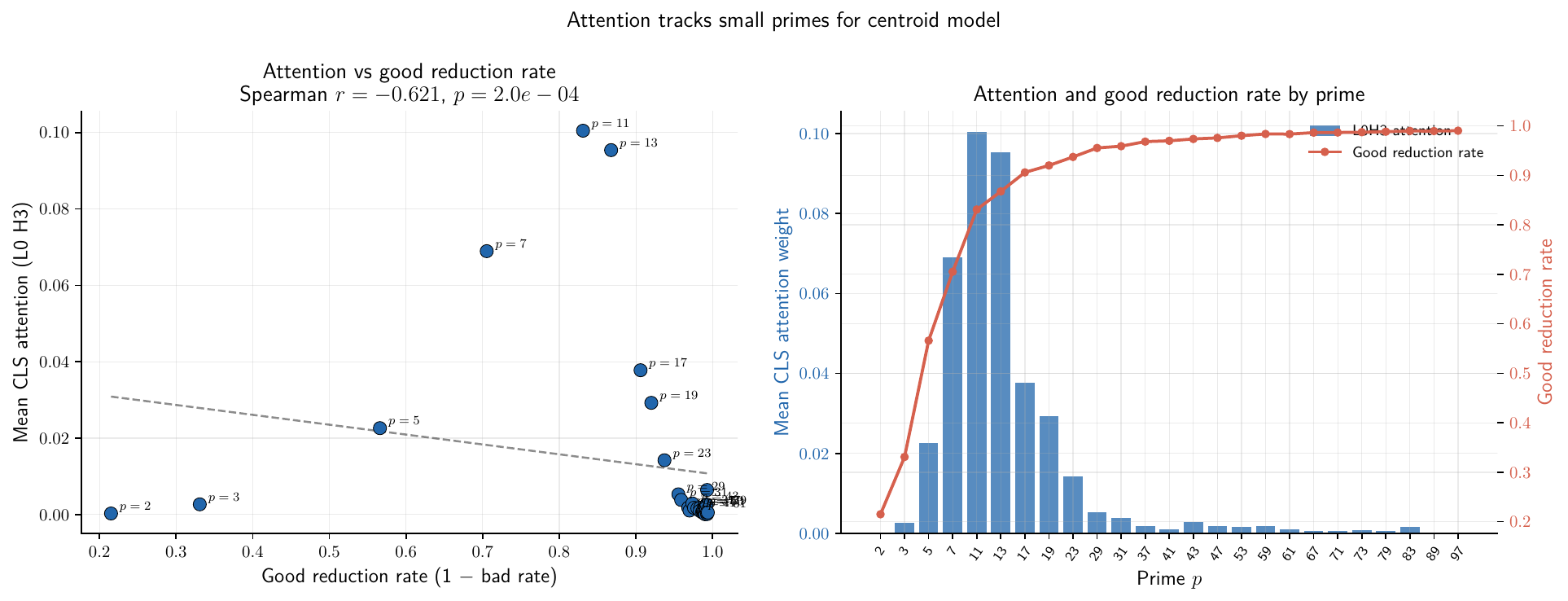}
  \caption{Note that the centroid model tracks small primes
  (in particular, $p=11, 13$) the most. This is not
explained by the rate of elliptic curves with good reduction
at $p$. For models trained on smaller trace sequence lengths,
confounding due to smaller number of primes was observed.}
  \centering
  \label{fig:fig2}
\end{figure}

\subsection{Attention Causality Dissociation}
Note that the centroid model concentrates attention on early
primes in several heads (visible in \cref{fig:fig2})
: in particular, the primes $p=11$
and $p=13$ receive the most attention across all layers.
However, from activation patching analysis, see
\cref{sec:activation_patching}. We see that the most
important prime in terms of causal patch effects is $p=31$,
followed by $p=13$. $p=11$ is not within the top ten most
causally impactful primes. Several composite positions also
have non-trivial causal effects. Refer to
\cref{tab:attn_vs_patch} for the top 10 positions in terms of
causal patch effects. This dissociation between
attention and causality is an instance of the finding of
Jain and Wallace \cite{jain2019attentionexplanation} in a
number-theoretic setting that attention analysis is not
necessarily reliable as a complete explanation of model
internals. In particular, attention analysis correctly gives
the coarse explanation that prime positions are important,
but does not explain which prime positions in particular are
the most significant. 

\begin{table}[h]
\centering
\caption{Attention vs.\ causal importance for the centroid model. The top-attended primes
($p=11, 13$) do not correspond to the most causally important positions under activation
patching. Notably, $p=11$ does not appear in the top 10 by causal effect, while $p=31$
(ranked first by patch effect) receives near-zero attention. Three composite positions
($a_9, a_{25}, a_{26}$) appear among the top 10 causal positions despite receiving
negligible attention.}
\label{tab:attn_vs_patch}
\begin{tabular}{clcc}
\toprule
\textbf{Rank} & \textbf{Position} & \textbf{Patch effect} &
\textbf{Attention (L0 H3)} \\
\midrule
1  & $a_{31}$ (prime)     & 0.034 & $<0.01$ \\ 
2  & $a_{13}$ (prime)     & 0.031 & 0.095 \\
3  & $a_{19}$ (prime)     & 0.022 & 0.029 \\
4  & $a_7$ (prime)        & 0.018 & 0.069 \\
5  & $a_{47}$ (prime)     & 0.018 & $<0.01$ \\
6  & $a_{17}$ (prime)     & 0.017 & 0.038 \\
7  & $a_9$ (composite)    & 0.016 & --- \\
8  & $a_{26}$ (composite) & 0.015 & --- \\
9  & $a_{25}$ (composite) & 0.015 & --- \\
10 & $a_5$ (prime)        & 0.014 & 0.022 \\
\midrule
--- & $a_{11}$ (prime)    & $<0.01$ & \textbf{0.101} \\
\bottomrule
\end{tabular}
\end{table}

\section{$L$-value Encoding}\label{sec:L_value}
\begin{figure}[htpb]
  \includegraphics[scale=0.5]{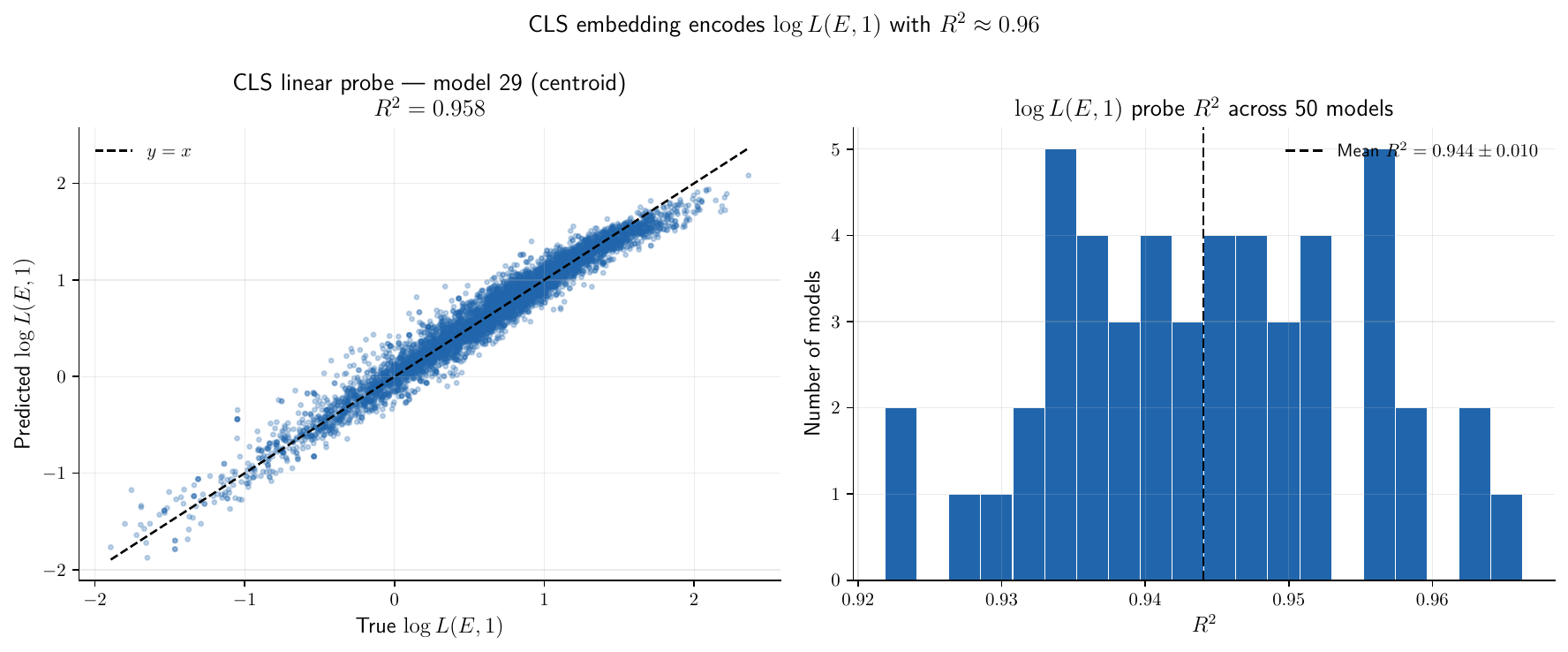}
  \caption{Scatter plot of CLS-predicted vs. true value of
  $\log{L(E, 1)}$ for the 50 trained models on the left
  and the explicit regression model on the right. Both
achieve $R^2$ values $> 0.94$.}
  \centering
  \label{fig:fig3}
\end{figure}

We fit a Ridge regression from the 128-dimensional CLS
embedding to $\log{L(E, 1)}$ for rank 0 test curves,
referenced against the LMFDB exact values of the
$L$-values. Across 50 runs, we observed an $R^2 = 0.944 \pm
0.011$. After controlling for the conductor (regressing out
$\log{N_E}$), we observe residual $R^2 = 0.962 \pm 0.011$.
We additionally trained a model to explicitly regress
$\log{L(E, 1)}$ from the same inputs, and this achieved an
$R^2 = 0.953$. This implies that the classification model
implicitly optimizes a near-complete $L$-value
approximation, as visible in \cref{fig:fig3}.

Probes for the remaining BSD-invariants recovered little to
no signal, as expected since global arithmetic invariants
are not determined by finitely many local traces. We defer
the fuller treatment (including a discussion of encoding the
real period $\Omega_E$) to the sequel.

\section{Activation Patching}\label{sec:activation_patching}

\subsection{Method}

We apply activation patching \cite{10.5555/3600270.3601532}
to identify which positions causally determine rank
prediction. In particular, for a clean rank 0 and a
corrupted rank 1 curve, we patch residual stream activation
$(l, p)$ from clean into corrupted and measure the
normalized logit difference: 
\[ \operatorname{patch}(l, p) =
  \frac{\Delta_{\mathrm{logit}}(\mathrm{patched}) -
  \Delta_{\mathrm{logit}}(\mathrm{corrupt})}{\Delta_{\mathrm{logit}}(\mathrm{clean})
    - \Delta_{\mathrm{logit}}(\mathrm{corrupt})}. \] 
Results are averaged over 200 pairs. We also note 
that the information flow structure was
qualitatively consistent across a sample of the 50
solutions.

\subsection{Direct Logit Attribution}

To quantify the relative contribution of the various model
components, we decompose the output logit difference into
contributions from each attention head and MLP layer by
direct logit attribution, following
\cite{elhage2021mathematical}. In particular, we have 
\[ \Delta_{\mathrm{logit}} = \sum_{\ell, h} \left (W_U \cdot
  z_{\mathrm{CLS}}^{(\ell, h)}\right ) + \sum_{\ell} \left
( W_U \cdot m_{\mathrm{CLS}}^{(\ell)} \right), \] where
$W_U$ is the unembedding matrix that stores the difference
direction of logit weights. The
left sum in the expression 
is the head contribution and the right sum is the
MLP contribution. 

Across 50 models, the layer-1 MLP dominates: its mean
absolute contribution is 3.2$\times$ larger than the layer 0
MLP and 7.5$\times$ larger than individual attention heads.
Attention heads collectively account for less than 15$\%$ of
total logit variance. This motivates the neuron-level
analysis of the layer-1 MLP in the next section. 

\begin{figure}[htpb]
  \includegraphics[scale=0.4]{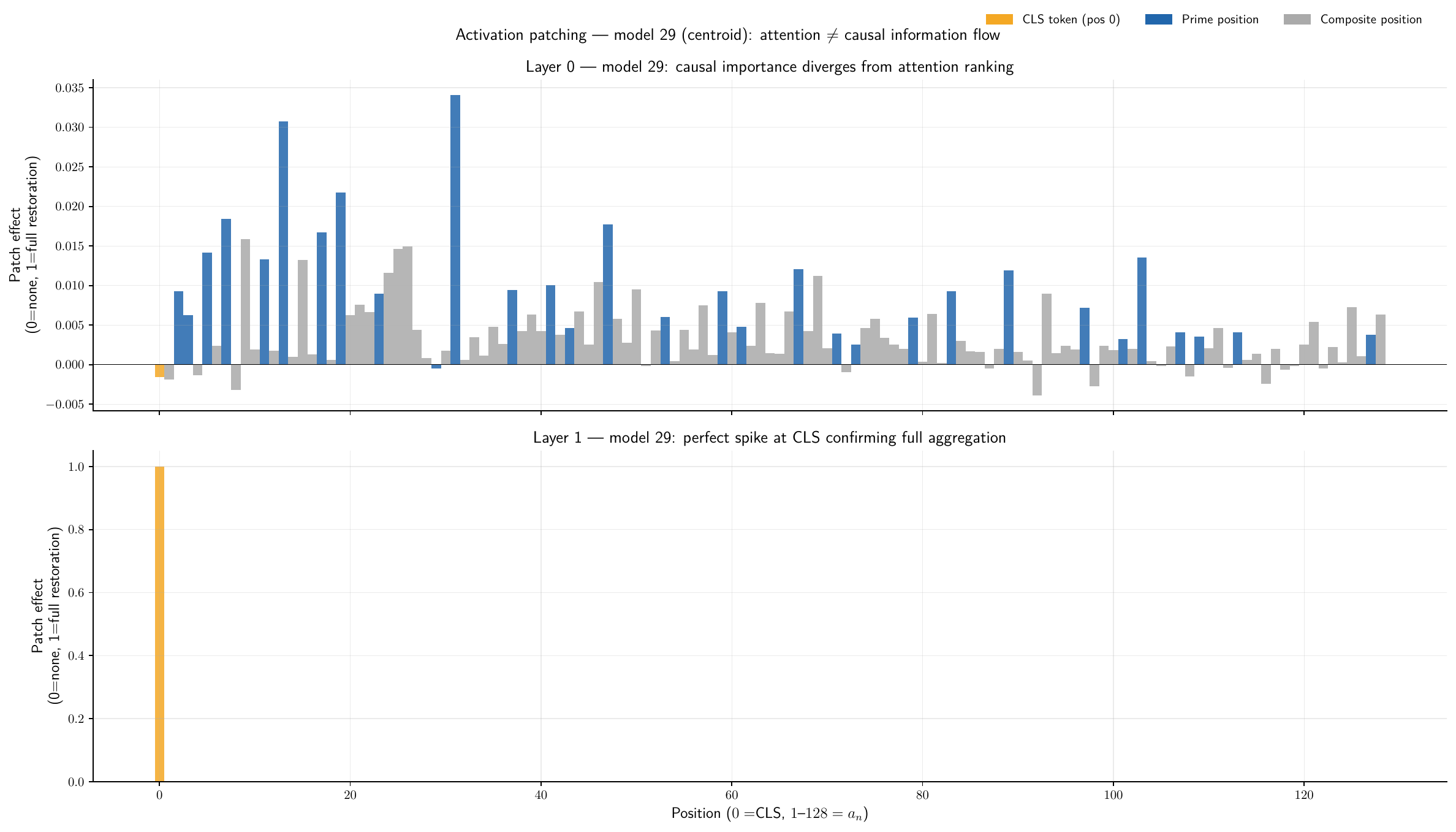}
  \caption{Two-panel activation patching figure for the
  centroid model. The top shows activation patching for
layer 0. Notice that the most significant primes are $p=31,
13, 19$, which differ from the primes that receive the most
attention by the model. The bottom shows the activation patching 
for layer 1, which has a perfect spike at the CLS position.}
  \centering
  \label{fig:fig4}
\end{figure}

\section{MLP Circuit Analysis and Mestre-Nagao Sums}
\subsection{Circuit Sparsity}
For each neuron $n$ in the layer 1 MLP (totalling 512), we
compute the Fisher discriminant score 
\[ F_n = \frac{|\overline{a_{n, 0}} - \overline{a_{n,
    1}}|}{\sqrt{\frac{\sigma_{n, 0}^2 + \sigma_{n,
  1}^2}{2}}}. \] Here, $\overline{a_{n, r}}$ is the mean
  post-ReLU activation for rank $r$ curves. We then fit a
  logistic probe from the top-$k$ neurons' activations and
  measure AUROC as $k$ increases. 

  We also for each $k$ select the top-$k$ neurons by
  $|w_n|$, where $w_n$ denotes the neuron's effective
  contribution to the logit difference under the model's
  \emph{own} weights. In particular, we have that the layer
  1 MLP output is 
  \[ \operatorname{MLP}(x) = W_2 
  \operatorname{ReLU}(W_1x + b_1) + b_2, \] 
  where $W_1 \in \mathbb R^{512 \times d}$ maps the CLS
  residual stream into the 512-dimensional hidden layer, and
  $W_2 \in \mathbb R^{d \times 512}$ maps back from the
  hidden layer to the CLS residual stream. The
  classification head computes $\Delta_{\mathrm{logit}} =
  v^{\top}c$ where $c$ is the CLS embedding and $v =
  w^{(0)} - w^{(1)}$ is the logit direction. The effective
  weight of a neuron $n$ is thus $w_n = (W_2^{\top}v)_n$,
  and the direct attribution score is 
  \[ \sum_{n \in S_k} w_nh_n, \] where $S_k$ is the set of
  analyzed neurons and $h_n$ is the post-ReLU activation
  of hidden neuron $n$. We then measure AUROC as $k$
  increases.

\begin{figure}[htpb!]
    \includegraphics[scale=0.5]{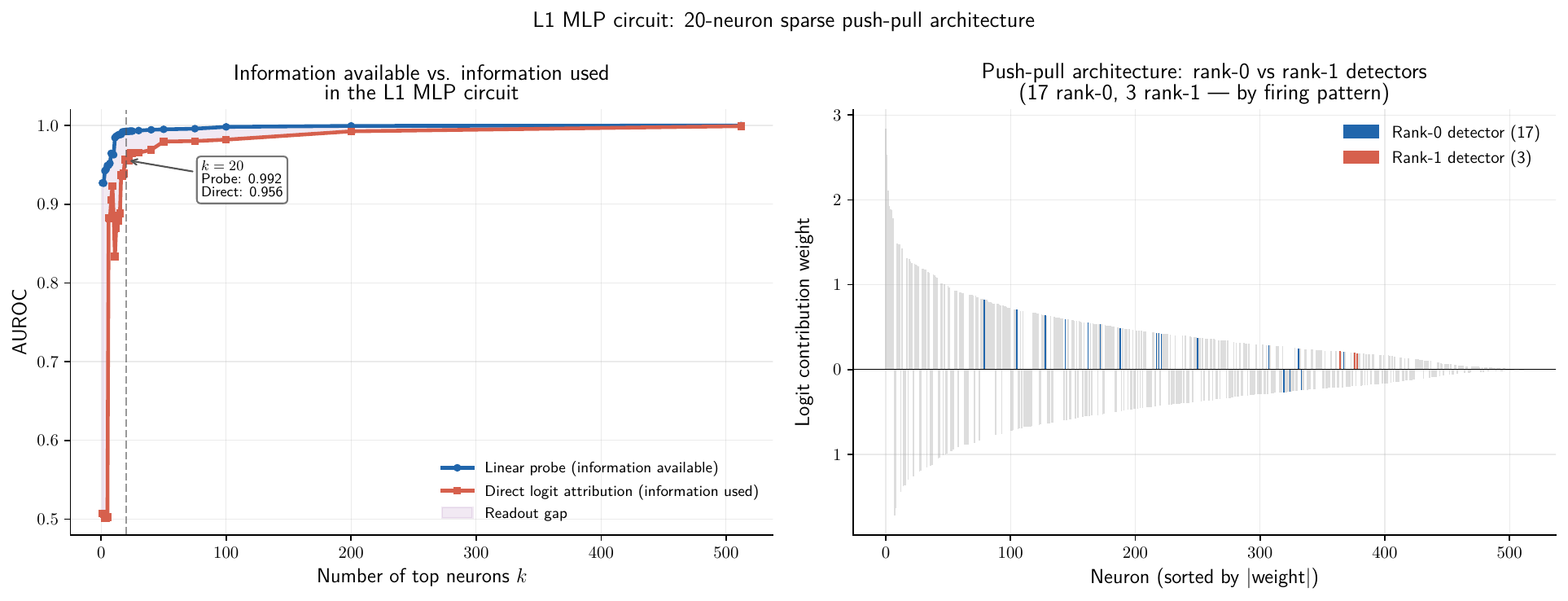}
    \caption{The left figure depicts the difference in AUROC
    vs. the number of top-$k$ neurons for both scores.
  Notice the divergence for low $k$ initially. The right
figure depicts logit contribution weights for all 512
neurons sorted by $|w_n|$, colored blue for rank 0 detector
neurons, and colored red for rank 1 detector neurons.}
    \centering
    \label{fig:fig5}
  \end{figure}

\begin{center}
  \begin{figure}[htpb]
    \includegraphics[scale=0.6]{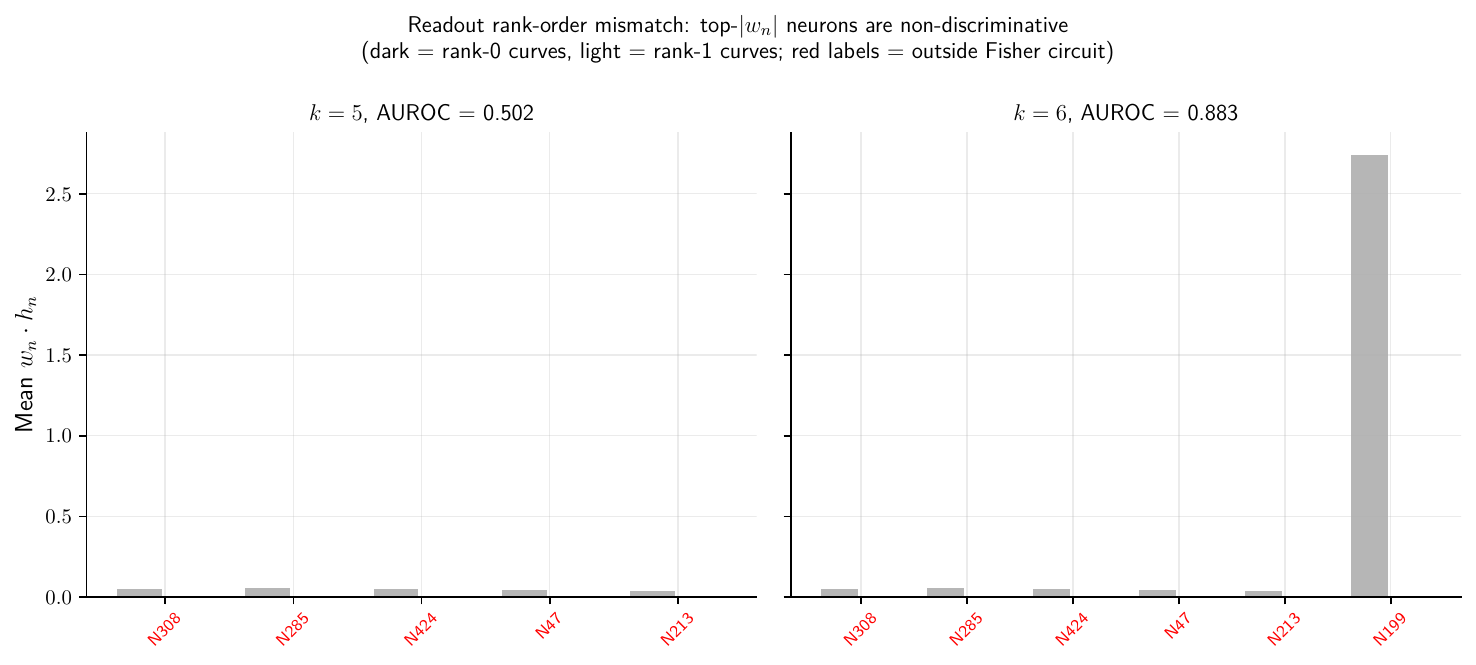}
    \caption{Note the jump in AUROC when the neuron N199 is
    added, despite it falling outside the top 20 neurons
  ordered by Fisher discriminant, indicative of sub-optimal
  readout.}
    \label{fig:interference}
  \end{figure}
  \end{center}

  The two scores diverge sharply initially before converging
  at $k=200$, see \cref{fig:fig5}. At the cutoff $k=20$, the
  linear probe achieves AUROC 0.992, while the direct logit
  attribution achieves AUROC 0.956, and the direct logit
  attribution curve is non-monotonic at small values of $k$.
  In particular, note that the direct logit attribution
  curve sits essentially at chance (0.501 - 0.507) for $k=1,
  \ldots, 5$ before jumping to AUROC 0.883 when the neuron
  N199 is included (refer to \cref{fig:interference} for a plot). 
  This reflects a rank-order mismatch
  between the readout circuit and discriminative circuit:
  the ten neurons with the largest readout magnitudes $|w_n|$ 
  all lie outside the Fisher top-100, and none belong to the
  20-neuron rank discriminative circuit: in particular,
  the model's 5 biggest readout weights
  point at neurons that are collectively inconsequential 
  to rank discrimination.

  The gap reflects ordering rather than orientation: a
  sign-correcting of the six circuit neurons whose readout
  sign disagreed with their firing pattern left the
  direct-attribution AUROC essentially unchanged at every $k$. 

  We classify each neuron by its \emph{firing pattern}: we
  compute $\Delta_n = \overline{a_{n, 0}} - \overline{a_{n, 1}}$ the
  mean post-ReLU activation differential. Neurons with
  $\Delta_n > 0$ are rank 0 detectors, and those with
  $\Delta_n < 0$ are rank 1 detectors. We classify
  explicitly by firing rather than by the sign of $w_n$ as
  they disagree for several circuit neurons.

  \subsection{Push-Pull Architecture}
  The 20 circuit neurons can be split by firing pattern into
  17 rank-0 detectors and 3 rank-1 detectors. Each neuron's
  pre-activation is well-approximated (with $R^2 = 0.81 -
  0.89$) by a prime-weighted linear form 
  \[ z_n \approx \sum_{p} c_p^{(n)}a_p + b_n, \] where $h_n
  = \operatorname{ReLU}(z_n)$. The regressions in
  \cref{sec:mn_sums} show that rank-0 detectors display
  coefficient profiles correlated with Mestre-Nagao weights,
  while rank-1 detectors display profiles anticorrelated
  with Mestre-Nagao weights (with Spearman $r$ ranging from
  $-0.48$ to $-0.56$). Each class thus fires on its own rank
  class and is nullified by the ReLU on the other class. 

  \begin{figure}[htpb]
    \includegraphics[scale=0.5]{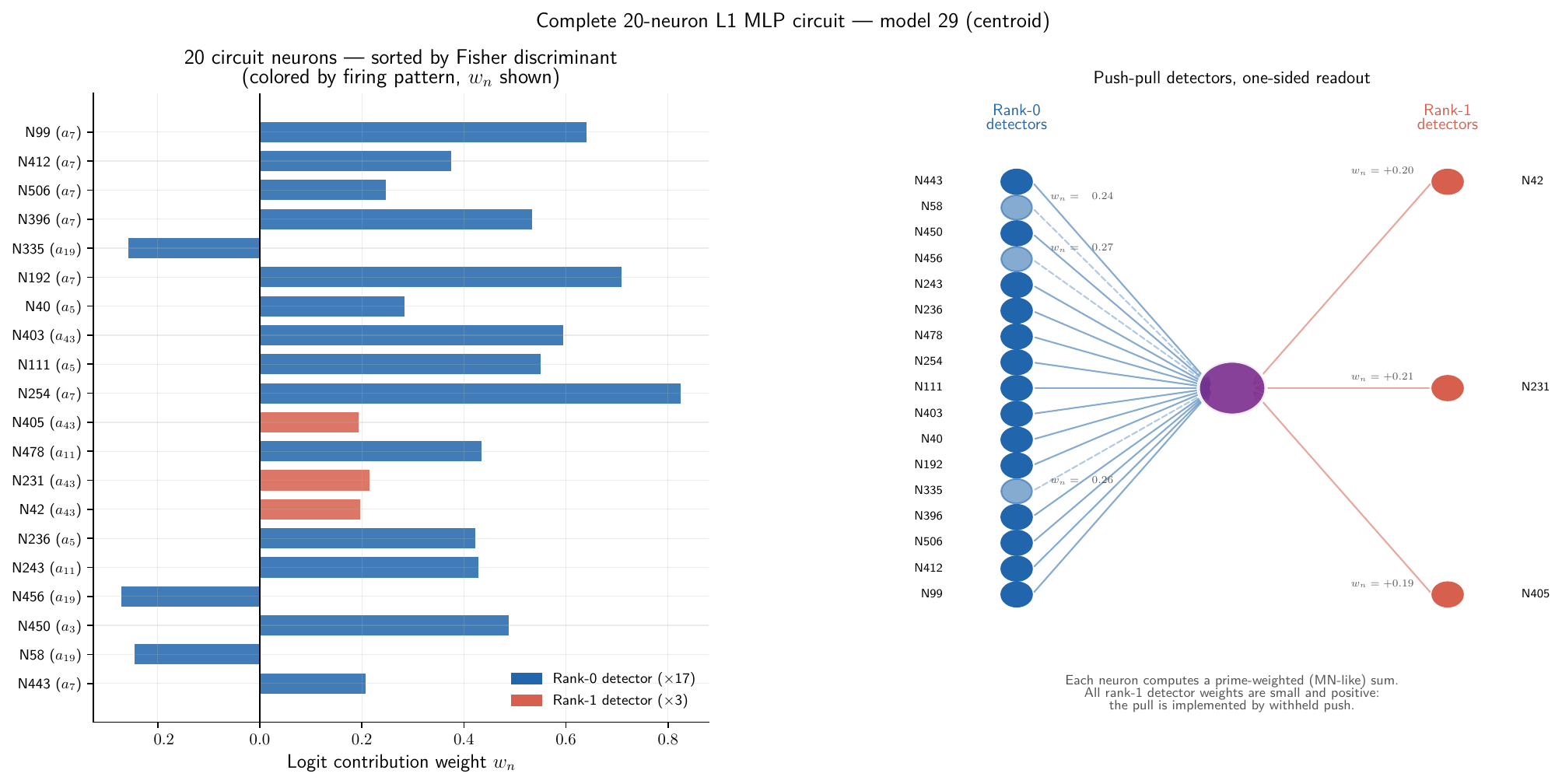}
    \caption{The left figure shows the 20 circuit neurons
    sorted by the Fisher discriminant plotted against their
  logit contribution weights and most-correlated primes. The
right figure shows the push-pull architecture of the
circuit: rank 0 and rank 1 detectors each compute a scaled
Mestre-Nagao partial sum.}
    \centering 
    \label{fig:circuit_summary}
  \end{figure}

  The logit weights then wire the detectors into a vote:
  this wiring is in particular strongly one-sided. Refer to
  \cref{fig:circuit_summary} for a plot of the circuit. 
  Firing
  class and the vote sign agree for 14 of the 20 circuit
  neurons: aligned rank-0 detectors push the rank-0 logit
  with weights up to $w_n = 0.83$. The six neurons that
  disagree are systematic: three rank-0 firing neurons N335,
  N456, and N58 (with $w_n = -0.26, -0.27, -0.24$
  respectively) push toward rank 1 when they fire, and the
  three rank-1 detectors N405, N231, and N42 (with $w_n =
  0.19, 0.21, 0.20$ respectively) push toward rank 0 when
  they fire. The rank-1 class does not receive a positive
  vote anywhere in the circuit. In particular we see that
  summing the mean signed contributions $w_nh_n$ over the
  circuit gives $9.71$ on rank-0 curves versus $1.19$ on
  rank-1 curves. The model signals rank-1 by the
  essential \emph{absence} of rank-0 push rather than by a
  pull in the negative direction. An ablation of N405
  shifted the mean logit difference by -0.026 on the curves
  where it fires, which shows that the positive weight
  convention is indeed correct. 

  Notice further that the misalignment is confined to small
  magnitudes as all misaligned neurons have $|w_n| \leq
  0.27$, whereas aligned detector neurons reach
  up to $|w_n| = 0.83$. The misalignment is also prime-structured: the
  three misaligned rank-0 detectors are all best correlated
  with $a_{19}$, and the three rank-1 detectors are best
  correlated with $a_{43}$.

  \subsection{Mestre-Nagao Sums}\label{sec:mn_sums}
\begin{figure}[htpb]
  \includegraphics[scale=0.4]{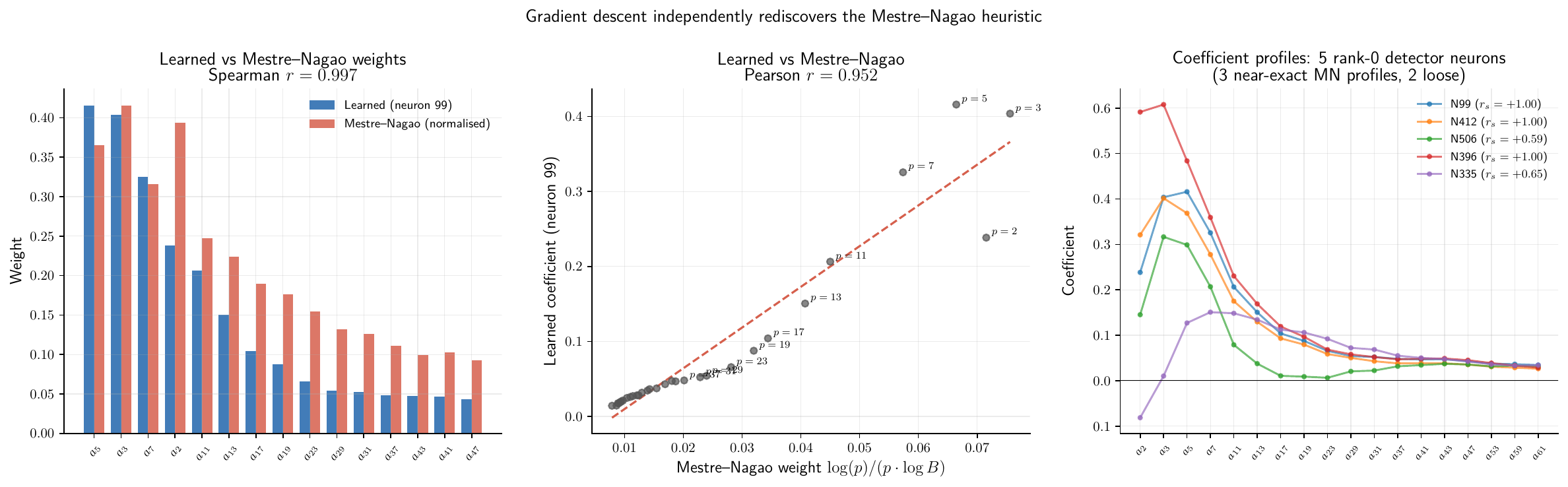}
  \caption{The left figure shows the learned neuron 99
  coefficients vs. the Mestre-Nagao weights at the top 15
primes by magnitude: note the near-identical profiles. The
center figure shows a scatter plot of learned coefficients
vs. Mestre-Nagao weight at each prime $p$. The right figure
shows coefficient profiles for 5 rank-0 detector neurons,
showing the parallel Mestre-Nagao-like structure across 3 of
the top 5 neurons in the circuit.}
  \centering
  \label{fig:fig6}
\end{figure}

In order to identify the learned input weighting, we fit a
Ridge regression from raw prime-indexed Frobenius traces
$\{a_p : p \leq 128\}$ to the pre-activation of each circuit
neuron. Of the top five discriminating rank-0 detectors,
three of them (N99, N412, and N396) have learned regression 
coefficients $\widehat{c_p}$ that closely
match the Mestre-Nagao weights $w_p =
\frac{\log{p}}{p\cdot \log{B}}$ from \cref{eqn:mn} 
(with Spearman $r \geq 0.997)$. The remaining
two neurons (N506 and N335) show the decaying shape more loosely
(with Spearman $r = 0.59, 0.65$ respectively). The circuit
computes parallel, approximately scaled copies of one
Mestre-Nagao partial sum rather than partitioning prime
ranges across the neurons. Refer to \cref{fig:fig6} for the
exact comparison of Mestre-Nagao weights and learned
coefficients for N99. 

\begin{figure}[htpb]
  \includegraphics[scale=0.4]{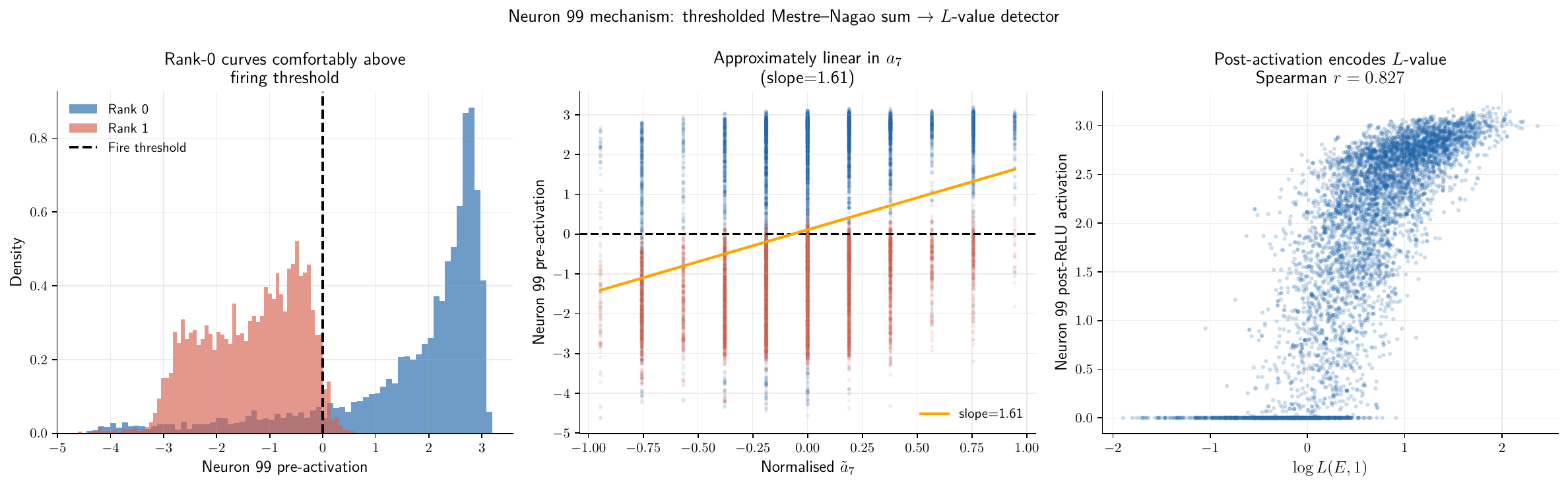}
  \caption{The left figure shows the pre-activation
  distributions for rank 0 vs. rank 1 curves. Note that the
two class distributions are cleanly 
separated by the firing threshold. The center
figure shows a scatter plot of pre-activation vs. the
Frobenius trace $a_7$, which shows a linear relationship.
The right figure shows the post-activation vs. $\log{L(E,
1)}$, which confirms that neuron 99 is an $L$-value
detector, as the Spearman $r = 0.827$.}
  \centering
  \label{fig:fig7}
\end{figure}

In \cref{fig:fig7}, we assemble the mechanism of N99. The
pre-activation distribution of the two rank classes is
well-separated about the firing threshold: rank-0 curves sit
comfortably above $z > 0$, and rank-1 curves below. The ReLU
activation itself acts as the decision boundary and the
neuron functions as a one-dimensional classifier on its
learned statistic. We note that said statistic is linear in
the traces, as seen in the center diagram, with slope 1.61
in $\tilde a_7$, the neuron's most correlated position. This
is consistent with the global linear fit of $R^2 = 0.89$. By
the Hasse bound, $a_7$ takes only the 11 values $-5, \ldots,
5$, which explains the vertical banding in the diagram. The
post-ReLU activation is a graded decision of rank: on rank-0
test curves, the post-activation increases monotonically
with the value of $\log{L(E, 1)}$ (with Spearman $r =
0.827$), so above threshold, the neuron's firing strength is
a proxy for the distance of $L(E, 1)$ from 0. Rank-1 curves
have $L(E, 1) = 0$ and are clipped to silence.

N99 thus implements a thresholded Mestre-Nagao partial sum:
a linear, Mestre-Nagao-weighted functional of the
prime-indexed traces whose ReLU output is simultaneously a
crude estimator of the $L$-value and a rank vote. The
sparse circuit aggregates twenty such similar detectors with
varying prime emphases and thresholds, and this ensemble
outperforms any single Mestre-Nagao sum (AUROC 0.95) and
explains at the neuron level the near-complete encoding of
$\log{L(E, 1)}$ in the CLS embedding observed in
\cref{sec:L_value}. 

\newpage
\printbibliography

@article{BirchSwinnertonDyer1965,
  author = {Birch, B. J. and Swinnerton-Dyer, P. F. A.},
  title = {Notes on elliptic curves. {II}},
  journal = {Journal f{\"u}r die reine und angewandte Mathematik},
  volume = {218},
  pages = {79--108},
  year = {1965},
  publisher = {De Gruyter}
}

@incollection{rubinstein2005computational,
  author    = {Rubinstein, Michael},
  title     = {Computational methods and experiments in analytic number theory},
  booktitle = {Recent Perspectives in Random Matrix Theory and Number Theory},
  editor    = {Mezzadri, F. and Snaith, N. C.},
  series    = {London Mathematical Society Lecture Note Series},
  volume    = {322},
  pages     = {425--506},
  publisher = {Cambridge University Press},
  year      = {2005}
}

@article{10.1016/j.jsc.2022.08.017,
author = {He, Yang-Hui and Lee, Kyu-Hwan and Oliver, Thomas},
title = {Machine learning invariants of arithmetic curves},
year = {2023},
issue_date = {Mar 2023},
publisher = {Academic Press, Inc.},
address = {USA},
volume = {115},
number = {C},
issn = {0747-7171},
url = {https://doi.org/10.1016/j.jsc.2022.08.017},
doi = {10.1016/j.jsc.2022.08.017},
journal = {J. Symb. Comput.},
month = mar,
pages = {478–491},
numpages = {14}
}

@article{kazalicki2023ranks,
  title   = {Ranks of elliptic curves and deep neural networks},
  author  = {Kazalicki, Matija and Vlah, Domagoj},
  journal = {Research in Number Theory},
  volume  = {9},
  number  = {3},
  pages   = {53},
  year    = {2023},
  publisher = {Springer},
  doi     = {10.1007/s40993-023-00462-w}
}

@misc{bieri2026murmurations,
  title         = {Murmurations, Mestre--Nagao sums, and Convolutional Neural Networks for elliptic curves}, 
  author        = {Joanna Bieri and Edgar Costa and Alyson Deines and Kyu-Hwan Lee and David Lowry-Duda and Thomas Oliver and Yidi Qi and Tamara Veenstra},
  year          = {2026},
  eprint        = {2603.17681},
  archivePrefix = {arXiv},
  primaryClass  = {math.NT}
}

@misc{babei2025learning,
  title         = {Learning Euler Factors of Elliptic Curves}, 
  author        = {Angelica Babei and Fran{\c{c}}ois Charton and Edgar Costa and Xiaoyu Huang and Kyu-Hwan Lee and David Lowry-Duda and Ashvni Narayanan and Alexey Pozdnyakov},
  year          = {2025},
  eprint        = {2502.10357},
  archivePrefix = {arXiv},
  primaryClass  = {math.NT}
}

@article{elhage2021mathematical,
  title={A Mathematical Framework for Transformer Circuits},
  author={Elhage, Nelson and Nanda, Neel and Olsson, Catherine and Henighan, Tom and Joseph, Nicholas and Mann, Ben and Askell, Amanda and Bai, Yuntao and Chen, Anna and Conerly, Tom and DasSarma, Nova and Drain, Dawn and Ganguli, Deep and Hatfield-Dodds, Zac and Hernandez, Danny and Jones, Andy and Kernion, Jackson and Lovitt, Liane and Ndousse, Kamal and Amodei, Dario and Brown, Tom and Clark, Jack and Kaplan, Jared and McCandlish, Sam and Olah, Chris},
  journal={Transformer Circuits Thread},
  year={2021},
  url={https://transformer-circuits.pub/2021/framework/index.html}
}

@misc{nanda2023progressmeasuresgrokkingmechanistic,
      title={Progress measures for grokking via mechanistic interpretability}, 
      author={Neel Nanda and Lawrence Chan and Tom Lieberum and Jess Smith and Jacob Steinhardt},
      year={2023},
      eprint={2301.05217},
      archivePrefix={arXiv},
      primaryClass={cs.LG},
      url={https://arxiv.org/abs/2301.05217}, 
}

@misc{alain2018understandingintermediatelayersusing,
      title={Understanding intermediate layers using linear classifier probes}, 
      author={Guillaume Alain and Yoshua Bengio},
      year={2018},
      eprint={1610.01644},
      archivePrefix={arXiv},
      primaryClass={stat.ML},
      url={https://arxiv.org/abs/1610.01644}, 
}

@inproceedings{10.5555/3600270.3601532,
author = {Meng, Kevin and Bau, David and Andonian, Alex and Belinkov, Yonatan},
title = {Locating and editing factual associations in GPT},
year = {2022},
isbn = {9781713871088},
publisher = {Curran Associates Inc.},
address = {Red Hook, NY, USA},
booktitle = {Proceedings of the 36th International Conference on Neural Information Processing Systems},
articleno = {1262},
numpages = {14},
location = {New Orleans, LA, USA},
series = {NIPS '22}
}

@book{diamond2005first,
  title={A First Course in Modular Forms},
  author={Diamond, Fred and Shurman, Jerry},
  isbn={978-0-387-23229-4},
  series={Graduate Texts in Mathematics},
  volume={228},
  year={2005},
  publisher={Springer},
  address={New York, NY},
  doi={10.1007/978-0-387-27226-9}
}

@inproceedings{devlin2019bert,
  title={BERT: Pre-training of Deep Bidirectional Transformers for Language Understanding},
  author={Devlin, Jacob and Chang, Ming-Wei and Lee, Kenton and Toutanova, Kristina},
  booktitle={Proceedings of the 2019 Conference of the North American Chapter of the Association for Computational Linguistics: Human Language Technologies, Volume 1 (Long and Short Papers)},
  pages={4171--4186},
  year={2019},
  organization={Association for Computational Linguistics}
}

@inproceedings{clark-etal-2019-bert,
    title = "What Does {BERT} Look at? An Analysis of {BERT}{'}s Attention",
    author = "Clark, Kevin  and
      Khandelwal, Urvashi  and
      Levy, Omer  and
      Manning, Christopher D.",
    editor = "Linzen, Tal  and
      Chrupa{\l}a, Grzegorz  and
      Belinkov, Yonatan  and
      Hupkes, Dieuwke",
    booktitle = "Proceedings of the 2019 ACL Workshop BlackboxNLP: Analyzing and Interpreting Neural Networks for NLP",
    month = aug,
    year = "2019",
    address = "Florence, Italy",
    publisher = "Association for Computational Linguistics",
    url = "https://aclanthology.org/W19-4828/",
    doi = "10.18653/v1/W19-4828",
    pages = "276--286"
}

@misc{lmfdb,
  shorthand = {LMFDB},
  author = {The {LMFDB Collaboration}},
  title = {The {L}-functions and modular forms database},
  howpublished = {\url{https://www.lmfdb.org}},
  year = {2026},
  note = {[Online; accessed 23 January 2026]}
}

@misc{dokchitser2002computingspecialvaluesmotivic,
      title={Computing special values of motivic L-functions}, 
      author={Tim Dokchitser},
      year={2002},
      eprint={math/0207280},
      archivePrefix={arXiv},
      primaryClass={math.NT},
      url={https://arxiv.org/abs/math/0207280}, 
}

@inproceedings{loshchilov2019adamw,
  title={Decoupled Weight Decay Regularization},
  author={Loshchilov, Ilya and Hutter, Frank},
  booktitle={International Conference on Learning Representations},
  year={2019},
  url={https://openreview.net/forum?id=Bkg6RiCqY7}
}

@article{ba2016layer,
  title={Layer normalization},
  author={Ba, Jimmy Lei and Kiros, Jamie Ryan and Hinton, Geoffrey E},
  journal={arXiv preprint arXiv:1607.06450},
  year={2016},
  url={https://arxiv.org}
}

@inproceedings{loshchilov2017sgdr,
  title={SGDR: Stochastic Gradient Descent with Warm Restarts},
  author={Loshchilov, Ilya and Hutter, Frank},
  booktitle={International Conference on Learning Representations},
  year={2017},
  url={https://openreview.net}
}

@misc{jain2019attentionexplanation,
      title={Attention is not Explanation}, 
      author={Sarthak Jain and Byron C. Wallace},
      year={2019},
      eprint={1902.10186},
      archivePrefix={arXiv},
      primaryClass={cs.CL},
      url={https://arxiv.org/abs/1902.10186}, 
}

@misc{elhage2022toymodelssuperposition,
      title={Toy Models of Superposition}, 
      author={Nelson Elhage and Tristan Hume and Catherine Olsson and Nicholas Schiefer and Tom Henighan and Shauna Kravec and Zac Hatfield-Dodds and Robert Lasenby and Dawn Drain and Carol Chen and Roger Grosse and Sam McCandlish and Jared Kaplan and Dario Amodei and Martin Wattenberg and Christopher Olah},
      year={2022},
      eprint={2209.10652},
      archivePrefix={arXiv},
      primaryClass={cs.LG},
      url={https://arxiv.org/abs/2209.10652}, 
}

@misc{goodfellow2015qualitativelycharacterizingneuralnetwork,
      title={Qualitatively characterizing neural network optimization problems}, 
      author={Ian J. Goodfellow and Oriol Vinyals and Andrew M. Saxe},
      year={2015},
      eprint={1412.6544},
      archivePrefix={arXiv},
      primaryClass={cs.NE},
      url={https://arxiv.org/abs/1412.6544}, 
}

@misc{garipov2018losssurfacesmodeconnectivity,
      title={Loss Surfaces, Mode Connectivity, and Fast Ensembling of DNNs}, 
      author={Timur Garipov and Pavel Izmailov and Dmitrii Podoprikhin and Dmitry Vetrov and Andrew Gordon Wilson},
      year={2018},
      eprint={1802.10026},
      archivePrefix={arXiv},
      primaryClass={stat.ML},
      url={https://arxiv.org/abs/1802.10026}, 
}

@misc{babei2026twistclassredundancydrives,
      title={How Twist Class Redundancy Drives the Prediction of Traces of Frobenius of Elliptic Curves}, 
      author={Angelica Babei and Ujjawal Shah and Malick Kebe},
      year={2026},
      eprint={2605.14288},
      archivePrefix={arXiv},
      primaryClass={math.NT},
      url={https://arxiv.org/abs/2605.14288}, 
}

@misc{entezari2022rolepermutationinvariancelinear,
      title={The Role of Permutation Invariance in Linear Mode Connectivity of Neural Networks}, 
      author={Rahim Entezari and Hanie Sedghi and Olga Saukh and Behnam Neyshabur},
      year={2022},
      eprint={2110.06296},
      archivePrefix={arXiv},
      primaryClass={cs.LG},
      url={https://arxiv.org/abs/2110.06296}, 
}

@misc{ainsworth2023gitrebasinmergingmodels,
      title={Git Re-Basin: Merging Models modulo Permutation Symmetries}, 
      author={Samuel K. Ainsworth and Jonathan Hayase and Siddhartha Srinivasa},
      year={2023},
      eprint={2209.04836},
      archivePrefix={arXiv},
      primaryClass={cs.LG},
      url={https://arxiv.org/abs/2209.04836}, 
}

@misc{kim2021birchswinnertondyerconjecturenagaos,
      title={From the Birch and Swinnerton-Dyer conjecture to Nagao's conjecture}, 
      author={Seoyoung Kim and M. Ram Murty},
      year={2021},
      eprint={2105.10805},
      archivePrefix={arXiv},
      primaryClass={math.NT},
      url={https://arxiv.org/abs/2105.10805}, 
}
\end{document}